\documentclass[letterpaper, 10 pt, conference]{ieeeconf}

\IEEEoverridecommandlockouts
\usepackage[style=ieee, backend=biber, sorting=none, citestyle=numeric]{biblatex}
\usepackage{flushend} 
\usepackage[section]{placeins} 
\usepackage{amsmath,amssymb,amsfonts}
\usepackage{algorithmic}
\usepackage[colorinlistoftodos,prependcaption,textsize=tiny]{todonotes}
\usepackage{graphicx}
\usepackage{tikz}
\usepackage{textcomp}
\usepackage{booktabs}
\usepackage{subfigure}
\usepackage{multirow}
\usepackage{svg}
\usepackage{makecell}
\usepackage{colortbl}
\usepackage{xcolor}
\usepackage{hyperref}
\usepackage{soul}
\usepackage{bbding}
\usepackage{subfigure}
\makeatletter
\let\labelindent\relax
\makeatother
\usepackage{enumitem}
\usepackage{placeins}
\usepackage{dblfloatfix}
\usepackage[ruled, noend, linesnumbered]{algorithm2e}
\def\BibTeX{{\rm B\kern-.05em{\sc i\kern-.025em b}\kern-.08em
    T\kern-.1667em\lower.7ex\hbox{E}\kern-.125emX}}

\newcommand{\figref}[1]{Fig.~\ref{#1}}
\newcommand{\tabref}[1]{Table~\ref{#1}}
\newcommand{\secref}[1]{Section~\ref{#1}}

\renewcommand{\eqref}[1]{Eq.~\ref{#1}}

\definecolor{somegray}{rgb}{0.5, 0.5, 0.5} 

\makeatletter
\newcommand*\titleheader[1]{\gdef\@titleheader{#1}}
\AtBeginDocument{%
  \let\old@title\@title
  \def\@title{%
    \vskip-2.5em 
    \bgroup\normalfont\large\centering\color{somegray}\@titleheader\par\egroup
    \vskip1em 
    \old@title}
}
\makeatother
\titleheader{\Large This paper has been accepted to ICRA 2025.}

\hypersetup{
    colorlinks=true,
    linkcolor=black, 
    filecolor=black, 
    urlcolor=gray,  
    citecolor=black
}

\title{\LARGE \bf Beyond Robustness: Learning Unknown Dynamic Load Adaptation for Quadruped Locomotion on Rough Terrain}
\author{Leixin Chang, Yuxuan Nai, Hua Chen and Liangjing Yang
\thanks{This work was supported by “Human Space X” Initiative Phase I: Tiantong Multidisciplinary Seed Grant from International Campus of Zhejiang University and in part by the Ministry of Education, China under the Industry-University Educational Collaboration Project under Grant 230904701283901.}
\thanks{Leixin Chang, Yuxuan Nai, Hua Chen, Liangjing Yang are with the
ZJU-UIUC Institute, Zhejiang University, Zhejiang, China (e-mail:
\{leixin.23, yuxuan.22, huachen, liangjingyang\}@intl.zju.edu.cn).}}

\begin{document}
\maketitle
\begin{abstract}
Unknown dynamic load carrying is one important practical application for quadruped robots. Such a problem is non-trivial, posing three major challenges in quadruped locomotion control. First, how to model or represent the dynamics of the load in a generic manner. Second, how to make the robot capture the dynamics without any external sensing. Third, how to enable the robot to interact with load handling the mutual effect and stabilizing the load. In this work, we propose a general load modeling approach called load characteristics modeling to capture the dynamics of the load. We integrate this proposed modeling technique and leverage recent advances in Reinforcement Learning (RL) based locomotion control to enable the robot to infer the dynamics of load movement and interact with the load indirectly to stabilize it and realize the sim-to-real deployment to verify its effectiveness in real scenarios. We conduct extensive comparative simulation experiments to validate the effectiveness and superiority of our proposed method. Results show that our method outperforms other methods in sudden load resistance, load stabilizing and locomotion with heavy load on rough terrain. 
\href{https://leixinjonaschang.github.io/leggedloadadapt.github.io/}{Project Page}.

\end{abstract}


\section{Introduction}
Thanks to extensive research in legged robots, quadruped robots manifest significant potential for applications in complex environments where traversing unstructured terrains like slopes, stairs, gravel, etc., is necessary \cite{biswal2021development, rudin2022learning, miki2024learning}. 
Among the various demands in such scenarios, load-carrying is one of the most significant and practical tasks. 
In load-carrying tasks, the most critical requirement is to maintain maneuverability and balance. 
Many practical scenarios require robots to carry or interact with dynamic loads, such as rolling or sliding loads. For example, a quadruped robot in delivery services could face situations where it needs to transport a heavy sphere-like object that could roll with the change of robot base pose. 
Generally, these dynamic loads can cause sudden shifts in the robot’s center of mass (CoM), leading to robot state variation, which in turn affects the load state.
The constantly changing nature of these loads and the interaction between dynamic loads and the robot present a significant challenge to maintaining balance and stability during movement, making dynamic load adaptation a critical area for further exploration. 

However, most previous research work handles the load from two perspectives. One of the perspectives is to view the effects of load as disturbance simplistically and resort to the robustness of the controller \cite{fey2024learning}, like domain randomization. The other is to make some restrictive assumptions on the load and develop the controller accordingly \cite{jin2021high}. Notably, both of these two major ways ignore the underlying dynamics of load states. 
\begin{figure}
    \centering    
    \includegraphics[width=\linewidth]{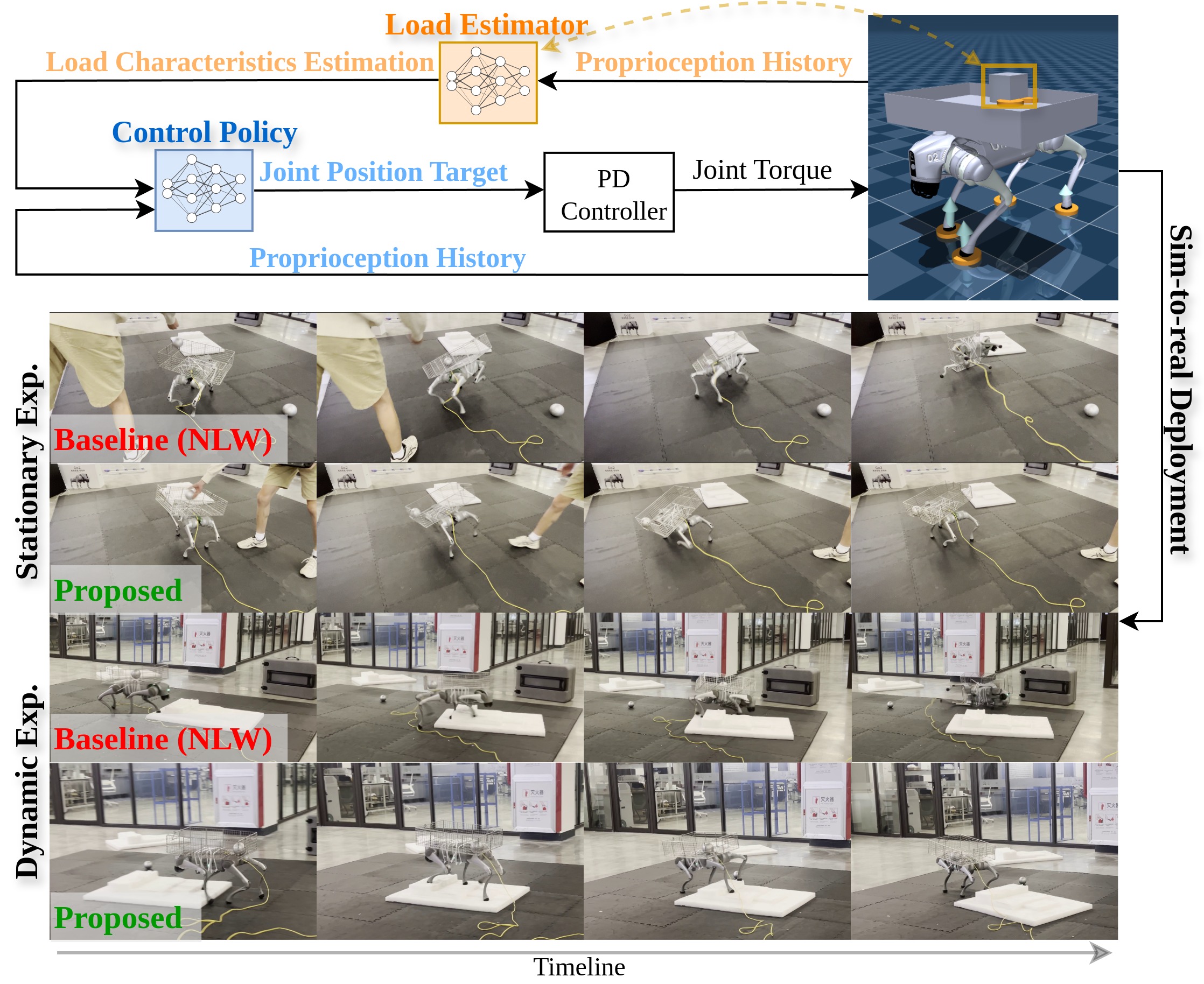}
    \caption{Learning general policy for quadrupedal unknown dynamic load adaption. The schematic diagram depicts the deployable policy trained by the proposed method. The first two columns and the second two columns of photos show the real deployment effect of Baseline (NLW) and the proposed method in stationary and dynamic tests.}
    \label{fig:cover_figure}
    \vspace{-15px}
\end{figure}
Starting from a new perspective that unifies capturing the load dynamics and the robot-load mutual effects, we propose a load modeling method that is general enough to help the robot learn a general policy for highly dynamic load handling.
We focus on making quadruped robot locomotion stable and maneuverable with the dynamic loads carried on while only employing the proprioceptive information without relying on any additional sensor data. Not only the robustness, our approach captures the dynamics of loads and realizes the adaptation to it. However, there are different forms in which dynamic loads can be mounted on the quadruped robot, such as pulling a cart with a rope, carrying loads suspended from a pole, or transporting dynamic objects like a watermelon placed in a crate mounted on the robot torso. The crate-based dynamic load scenario reflects real-world applications, where animals like dogs or horses are already used to carry loads that shift and affect their balance. Its practical relevance and the complexity of handling such dynamic interactions make it an ideal case for study, so we limit our target scenario to the dynamic load on the crate mounted on the robot base.

Our main contributions are summarized as follows. First, we propose a general modeling method for the external load that leads to straightforward external disturbance resistance and load adaptation, which is called \textbf{load characteristics modeling}. Second, 
through utilizing the load characteristics modeling, we propose an RL policy training framework for unknown dynamic load adaptation enabling the robot to adapt to dynamic load while locomoting on the rough terrain with only proprioception provided. Third, we conduct extensive comparative simulation experiments and physical experiments to validate the effectiveness of our proposed general load modeling method and policy training framework.

\secref{sec:related_work} introduces the representative work related to the legged locomotion with loads. 
\secref{sec:method} formulates the problem and introduces the proposed framework. 
\secref{sec:experiments} displays the results of conducted simulation experiments and sim-to-real experiments on the physical quadruped robot. 
Finally, \secref{sec:conclusion_future} concludes this work and briefly discusses directions for future work.

\section{Related Work} \label{sec:related_work}

\subsection{Reinforcement Learning for Quadruped Locomotion}
Recently, Reinforcement Learning (RL) has achieved impressive success in quadruped locomotion control, showing great generalizability and robustness in handling challenging terrains\cite{hwangbo2019learning, lee2020learning, nahrendra2023dreamwaq, margolis2024rapid, wu2023learning}. 
Teacher-student architecture effectively leverages information accessible exclusively in simulation to improve robot training and enables effective sim-to-real deployment with limited proprioception information\cite{lee2020learning, kumar2021rma, wang2024combining}. A recent work, \textit{DreamWaQ}\cite{nahrendra2023dreamwaq}, proposed an asymmetric actor-critic framework\cite{pinto2017asymmetric} where the actor receives partial observations, while the critic has access to full state information, enabling the policy to learn robust locomotion strategies by leveraging privileged information during training.
In essence, this method correlates the distribution of proprioception observation and that of privileged observation like terrain information, enabling the robot to implicitly imagine the environment information. Different from these two implicit ways, \cite{ji2022concurrent} proposed a concurrent training pipeline to train a control policy and an explicit state estimator for unobservable states. 
The control policy is trained via Proximal Policy Optimization (PPO)\cite{schulman2017proximal} and the estimator is trained via supervised learning using ground truth from the simulator as the reference. Notably, the concurrency of training makes the control policy network adapt to the performance of the estimator network, which can be seen as a form of backward compatibility. 
In this paper, we leveraged the concurrent training technique and the asymmetric actor-critic architecture to train a policy and load estimator that can make the robot adapt to the dynamic unsensed load and stabilize the load while locomotion on rough terrain with only proprioceptive observation.

\subsection{Load Adaptation}
In this work, we focus on a quadruped robot's adaptation for unknown dynamic loads, which means that only the proprioceptive information is accessible when the robot handles the impacts of dynamic load in locomotion with the load. 
\cite{liu2023load} introduces an observer to estimate the ground reactive force and subsequently identify the position and mass of the unknown load on the robot with only proprioceptive sensing. \cite{sombolestan2021adaptive} introduced an adaptive force-based control framework where the load, rigidly attached to the trunk, was treated as the uncertainty of robot trunk mass and Center of Mass (CoM). 
\cite{jin2021high} takes the static load adaptation as an online parameter identification problem. All of these three representative model-based methods mentioned above can only handle the static load, ignoring the mutual effects of load and robot dynamics. 
\cite{dao2022sim} explores the use of RL for bipedal locomotion control under unknown dynamic loads without additional sensing. 
In the training, the dynamic load is directly added to the robot without any specific design for load handling except for domain randomization, where the load is just viewed as a source of disturbance and the policy can not infer and then adapt to the dynamics of the load. Finally, all of the works mentioned above did not consider rough terrains like slopes, gravel and even steps, which are highly likely to be encountered in the real world.

\section{Method} \label{sec:method}

\begin{figure*}[htb!]
    \centering
    \includegraphics[width=0.85\linewidth]{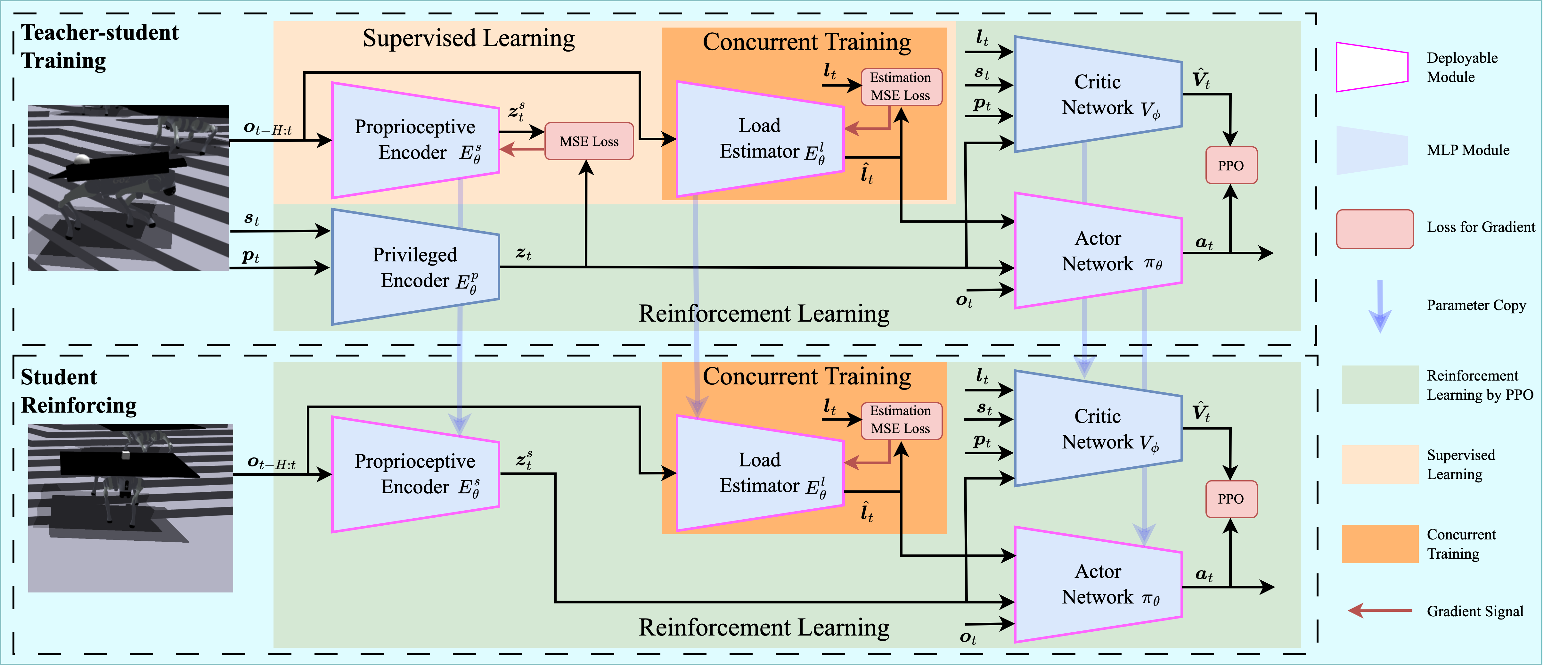}
    \caption{Overview of the proposed framework for dynamic load adaptation on the quadruped robot.} 
    \label{fig:training_pipeline}
    \vspace{-10px}
\end{figure*}

Our goal is to develop an RL-control framework for a quadruped robot that can handle the sustained and varying external forces caused by the highly unknown dynamic loads and stabilize the load on the robot with no external sensing information but only the proprioception from the Inertial Measurement Unit (IMU) and joint encoders. 
In this section, we elaborate on how we leverage our proposed general load modeling method, combined with the teacher-student training architecture and concurrent training framework to address the challenge of dynamic load adaptation.

\subsection{General Load Modeling}
The key to enabling quadruped robots to adapt to the unknown dynamic load lies in making the robot capture the dynamics of the load movements for further capturing the dynamics of the external wrench resulting from the load.
The first question is how to represent the load and the dynamics in a form that is general enough so that the dynamic load can be generally incorporated into the learning framework.
Here, we introduce \textbf{load characteristics}, comprising the load's position, velocity (in the robot frame), mass, and friction coefficient.
Load mass can make the total mass of the system change and the load position determines how much the system CoM shifts given a certain load mass, both of which influence the system dynamics significantly.
Load velocity and the friction coefficient determine how fast this kind of influence on system dynamics varies.
Therefore, we model the dynamic load as load characteristics composed of the load's mass, friction coefficient, position, and velocity. 
This effectiveness of load characteristics modeling can refer to the performance of Oracle model \secref{sec:experiments}.

\subsection{Proposed Training Framework}
An overview of the proposed framework for dynamic load adaptation is illustrated in \figref{fig:training_pipeline}.
Briefly, we employed the proposed load characteristics modeling with the teacher-student and concurrent training technique and asymmetric actor-critic architecture to train a policy that can control the robot to adapt to unknown dynamic load in the locomotion on rough terrains
given partially observable robot states. 

In the following, we denote  $\boldsymbol{o}_t\in\mathbb{R}^{45}$ including robot angular velocity, gravity vector, velocity command in the robot base frame, joint positions and velocities, and the previous action as the proprioception and  $\boldsymbol{o}_{t-H:t}$ as its history where $H$ represents the history length. $\boldsymbol{s}_t\in\mathbb{R}^{264}$ consists of $\boldsymbol{o}_t$, linear velocity, terrain heights sample, joint torques, joint accelerations and feet contact forces. 
$\boldsymbol{p}_t\in\mathbb{R}^{62}$ denotes the dynamics randomization parameters including randomization for PD control gains, motor strength, leg mass, and CoM, base mass, and CoM. 
$\boldsymbol{z}_t\in\mathbb{R}^{32}$ is the compressed latent representation of the full robot states $\boldsymbol{s}_t$ and dynamics randomization parameters $\boldsymbol{p}_t$ and $\boldsymbol{z}^{s}_t\in\mathbb{R}^{32}$ is the reconstructed latent representation of $\boldsymbol{z}_t$.
$\boldsymbol{l}_t\in\mathbb{R}^{8}$ represents the load characteristics obtained from simulation, composed of $[\text{load pos., load vel., load mass, load fric. coef.}]$. 
And $\hat{\boldsymbol{l}}_t\in\mathbb{R}^{8}$ denoted the estimated load characteristics. 
$\boldsymbol{a}_t$ is the action output from policy and $\hat{V}_t$ refers to the value estimation by the critic network. 

The whole training process is divided into two phases: teacher-student training phase and student reinforcing phase. 

\subsubsection{Teacher-student Training}
In the teacher-student training phase, the privileged encoder $E_{\theta}^p$ compresses the full robot states $\boldsymbol{s}_t$ and dynamics randomization parameters into the latent vector $\boldsymbol{z}_t$. The latent representation $\boldsymbol{z}_t$ is mapped onto a unit hypersphere with L2-normalization. Actor network $\pi_{\theta}$ takes $\boldsymbol{z}_t$, proprioception $\boldsymbol{o}_t$ and estimated load characteristics $\hat{\boldsymbol{l}}_t$ as input and the critic $V_{\phi}$ take $\boldsymbol{s}_t$, $\boldsymbol{z}_t$, load characteristics ground truth $\boldsymbol{l}_t$, dynamics randomization $\boldsymbol{p}_t$ as input. $E_{\theta}^p$, $\pi_{\theta}$ and $V_{\phi}$ undergo the RL training with PPO algorithm. 
To make the proprioceptive encoder $E_{\theta}^s$ able to reconstruct the compressed robot states and environment information for better locomotion performance with only proprioception, the reconstruction loss is introduced to update the parameter of the proprioceptive encoder by minimizing the difference of $\boldsymbol{z}_t$ and $\boldsymbol{z}_t^s$, and the Monte-Carlo approximation of the reconstruction loss is defined in \eqref{eq:proprio_loss}, 
\begin{equation}
    L^{\mathrm{rec}}(\theta)= \frac{1}{|\mathcal{D}|T}\sum_{\tau\in\mathcal{D}} \sum_{t=0}^{T} \left\| \boldsymbol{z}_t^s - \boldsymbol{z}_t \right\|_{2}^{2}
    \label{eq:proprio_loss}
\end{equation}
where the $\mathcal{D}$ is the collected trajectory by running the policy $\pi_\theta$ and $T$ is the length of corresponding trajectory $\tau$. We train the proprioceptive encoder $E_{\theta}^s$ with loss $ L^{\mathrm{rec}}(\theta)$ as a supervision signal 
For a limited condition that the load is unsensed, we train a load characteristics estimator $E_{\theta}^l$ in a concurrent training manner with load estimation loss. The Monte-Carlo approximation of load estimation loss is defined as \eqref{eq:load_loss}, where weight $\boldsymbol{w}_l$ is a hyperparameter and introduced to emphasize certain aspects of the estimation error.
\begin{equation}
\begin{aligned}
    L^{\text{est}}(\theta) = \frac{1}{|\mathcal{D}|T} \sum_{\tau \in \mathcal{D}} \sum_{t=0}^{T} \left\| \boldsymbol{w}_l \odot (\hat{\boldsymbol{l}}_t - \boldsymbol{l}_t) \right\|_2^2
\end{aligned}
    \label{eq:load_loss}
\end{equation}

\subsubsection{Student Reinforcing}
In the teacher-student training phase, the proprioceptive encoder $E_{\theta}^s$ undergoes supervised training trying to imitate the privileged encoder. But there is always a slight difference between the $\boldsymbol{z}_t^s$ and $\boldsymbol{z}_t$. This difference causes a deviation between the actions output by the student policy and those output by the teacher policy, leading to a degradation in locomotion performance. Therefore, we reinforce the student policy with PPO in the same simulation environment after the teacher-student training phase to finetune both the proprioceptive encoder and actor network with reward signals. 

\subsection{Reward Design}
\tabref{tab:reward} details of the reward terms employed. Most locomotion-related reward functions are shaped by referring to \cite{rudin2022learning, wang2024combining}. 
The term designed specifically for dynamic load adaptation is the load linear velocity term. $\boldsymbol{v}_\text{load}$ represents the load velocity in the robot base frame. This term encourages the robot to manipulate the dynamic load into a static state relative to the robot. 
Notably, there's no specific orientation term to keep the base flat. However, as shown in \secref{sec:experiments}, the trained policy naturally keeps the base flat to stabilize the load, which is an emergent behavior learned through training.

\begin{table}[h!]
\vspace{-5px}
\centering
\caption{Reward Terms}
\begin{tabular}{llr}
\toprule
\textbf{Reward Term}           & \textbf{Expression}    & \textbf{Weight}     \\ 
\midrule
Lin. velocity tracking         & $\exp\left(-4 \Vert v^\text{cmd}_{xy} - v_{xy}\Vert^2\right)$   & $2.0$\\ 
Ang. velocity tracking         & $\exp\left(-4 (\omega^\text{cmd}_z - \omega_z)^2\right)$        & $0.5$\\ 
Lin. velocity (z)              & $v^2_z$                                                    & $-2.0$\\ 
Ang. velocity (xy)             & $\Vert\omega_{xy}\Vert^2_2$                                & $-0.05$\\ 
Joint acceleration             & $\Vert\ddot{q}\Vert^2_2$                                    & $-2.5 \times 10^{-7}$ \\ 
Joint power                    & $\vert\tau\vert \vert\dot{q}\vert^T$                             & $-2 \times 10^{-5}$\\ 
Joint torque                   & $\Vert\tau\Vert^2_2$                                       & $-1 \times 10^{-5}$\\ 
Base height                    & $(h^\text{des} - h)^2$                                          & $-1.2$\\ 
Action rate                    & $\Vert a_t - a_{t-1}\Vert^2_2$                             & $-0.02$\\ 
Action smoothness              & $\Vert a_t - 2a_{t-1} - a_{t-2}\Vert^2_2$                  & $-0.001$\\ 
Collision                      & $n_\text{collision}$                                            & $-1.0$\\ 
Joint limit                    & $n_\text{limitation}$                                           & $-2.0$\\ 
Feet air time                  & $r_{ft}$                                                   & $1.0$\\ 
Feet contact forces            & $r_{fcf}$                                                  & $-2.0$\\ 
\textbf{Load lin. velocity}             & $ \frac{1}{1 + \boldsymbol{v}_\text{load}}$                   & $2.0$\\ 
\bottomrule
\end{tabular}
\label{tab:reward}
\vspace{-10px}
\end{table}

\subsection{Details of Training Setting} \label{sec:training_details}
\subsubsection{Network Architecture}
All the network modules illustrated in \figref{fig:training_pipeline} are designed as the Multi-Layer Perceptron (MLP) with Exponential Linear Unit (ELU) activation. The details of the networks can be found in \tabref{tab:network_architecture}.
\begin{table}[h!]
\vspace{-5px}
\centering
\caption{Network Architectures}
\setlength{\tabcolsep}{3pt} 
\begin{tabular}{lcccc}
\toprule
\textbf{Module} & \textbf{Notation} & \textbf{Inputs} & \textbf{Hidden Layer} & \textbf{Outputs} \\ 
\midrule
Privileged Enc. & $E^{p}_{\theta}$ & $\boldsymbol{s}_{t}$, $\boldsymbol{p}_t$ & [512, 256, 128] & $\boldsymbol{z}_{t}$ \\
Proprio. Enc. & $E^{s}_{\theta}$ & $\boldsymbol{o}_{t-H:t}$ & [512, 256, 128] & $\boldsymbol{z}^{s}_{t}$ \\ 
Load. Est. & $E^{l}_{\theta}$ & $\boldsymbol{o}_{t-H:t}$ & [512, 256, 64] & $\hat{\boldsymbol{l}}_{t}$ \\ 
Actor & $\pi_{\theta}$ & $\boldsymbol{l}_{t}, \boldsymbol{o}_{t}, \boldsymbol{z}_{t}(\boldsymbol{z}^s_{t})$ & [512, 256, 128] & $\boldsymbol{a}_{t}$ \\ 
Critic & $V_{\phi}$ & $\boldsymbol{l}_{t}, \boldsymbol{s}_{t}, \boldsymbol{p}_{t}, \boldsymbol{z}_{t}(\boldsymbol{z}^s_{t})$ & [512, 256, 128] & $\hat{V}_{t}$ \\ 
\bottomrule
\end{tabular}
\label{tab:network_architecture}
\end{table}

\subsubsection{Simulated Training Environment}
We use the IssacGym simulator\cite{makoviychuk2021isaac} to build the simulation environment for training and train 8192 agents in parallel. Following \cite{rudin2022learning}, we set the maximum time duration of an episode as 20 seconds, corresponding to 1000 time steps with a control and state sampling frequency of 50 Hz. 
\begin{figure}
    \centering
    \includegraphics[width=0.9\linewidth]{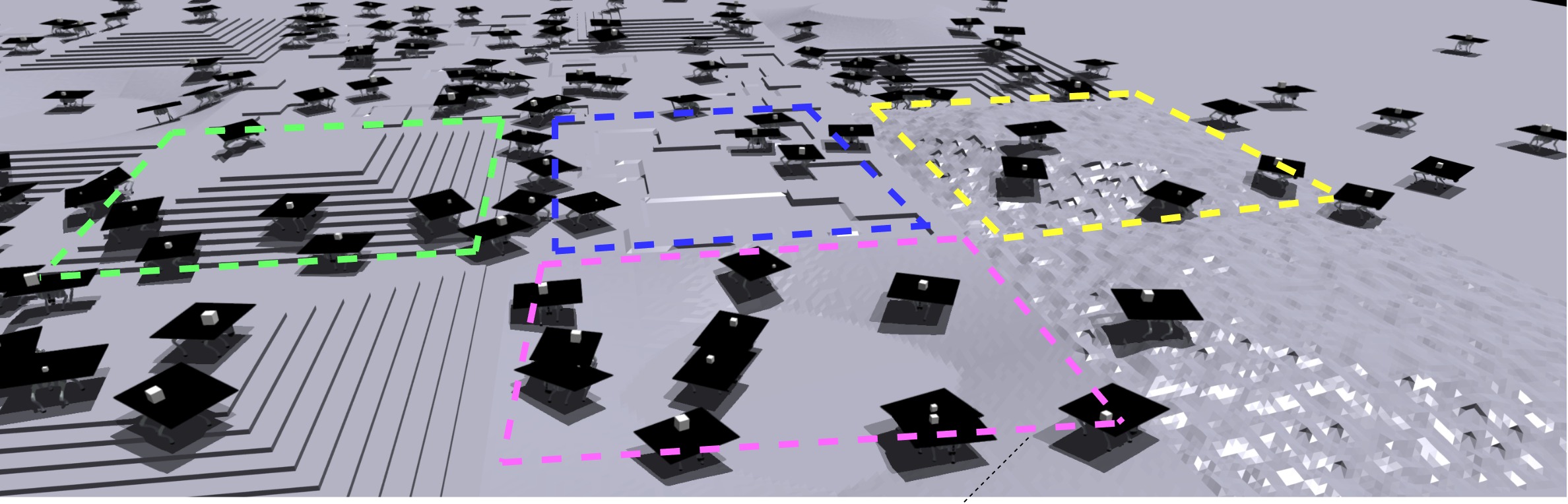}
    \caption{Terrain types in the training. Green: stairs, blue: discrete obstacles, yellow: rough terrains, pink: smooth slope.}
    \label{fig:terrain_type}
    \vspace{-15px}
\end{figure}
\begin{table}[htb!]
    \centering
    \caption{Domain Randomization Setting}
    \begin{tabular}{lll}
    \toprule
         \textbf{Randomization Term} & \textbf{Range}  & \textbf{Unit} \\
    \midrule
    Link mass    & $[0.8, 1.2]$ $\times$ nominal value   & Kg   \\
    Payload mass & $[-1, 3]$  & Kg \\
    CoM of base  & $[-5, 5]^3$  & cm   \\
    CoM of leg   & $[-1.5, 1.5]^3$  & cm  \\
    Friction     & $[0.05, 1.25]$  & - \\
    Joint $K_p$ factor  & $[0.8, 1.2]$   & N$\cdot$rad   \\
    Joint $K_d$ factor  & $[0.8, 1.2]$    & N$\cdot$rad/s \\
    Motor strength factor & $[0.8, 1.2]$   &    \\
    Action delay  & $[0, 10]$   & ms   \\ 
    Load mass & $[0.001, 8]$ & Kg \\
    Load size & $[0.025, 0.15]$& m \\
    Load fric. coef. & $[0.001, 0.2]$&  \\
    Load init. vel. & $[0, 0.5]$ & m/s  \\
    \bottomrule
    \end{tabular}
    \label{tab:domain_randomization}
    \vspace{-10px}
\end{table}
\begin{table}[htb!]
\centering
\caption{Hyper Parameters for Training}
\begin{tabular}{ll}
\toprule
\textbf{PPO Training} & \\ 
\midrule
Batch size           & 8192 $\times$ 24 \\ 
Mini-batch size      & 8192 $\times$ 6  \\ 
Number of epochs     & 5   \\ 
Clip range           & 0.2 \\
Entropy coefficient  & 0.01 \\ 
Discount factor      & 0.99 \\ 
GAE discount factor  & 0.95 \\ 
Desired KL-divergence  & 0.01 \\ 
Learning rate         & adaptive  \\ 
Teacher-student iteration & 7500 \\
Student reinforce iteration & 1500\\
\toprule
\textbf{Training for Proprio. Enc. \& Load Est.}  &  \\ 
\midrule
Batch size   & 8192 $\times$ 24 \\ 
Mini-batch size   & 8192 $\times$ 6 \\ 
Number of epochs  & 5  \\ 
Learning rate     & 1 $\times$ 10$^{-3}$ \\ 
Loss weight of $L^{est}_{\theta}$ &  $[3,3,3,1,1,1,10,10]$ \\
\bottomrule
\end{tabular}
\label{tab:training_parameter}
\vspace{-10px}
\end{table}
The episode is terminated upon the robot falling, timeout, or load falling. The joint PD controller parameters are set to be $kp = 20.0$, $kd = 0.5$ for the Unitree Go2 model. The length $H$ of the observation history $\boldsymbol{o}_{t-H:t}$ is 15 and the algorithm performed an iteration every 24 timesteps, following the setting in\cite{margolis2024rapid}. Hyperparameters for RL training and supervised training are displayed in \tabref{tab:training_parameter}. And we take the terrain curriculum strategy similar to \cite{margolis2024rapid} and the terrain types are shown in \figref{fig:terrain_type}. The maximum height of the step and the discrete obstacles are 8.5 cm and 9.5 cm, respectively. During the training, linear velocity commands and angular velocity commands are uniformly and randomly sampled from a range $[1, -1]$ m/s and $[1, -1]$ rad/s, respectively. 
To simulate the impact of dynamic load, we attach a $0.6$ m $\times$ $0.8$ m plate to the robot base and add a dynamic cube load with initial position, initial velocity, mass, size, friction coefficient and initial velocity randomized, which can refer to \tabref{tab:domain_randomization}.

To narrow the sim-to-real gap, we perform the randomization for the robot base mass and CoM, robot leg mass and CoM, friction coefficient between the rigid bodies and ground, PD gains, motor strength and action delay. The parameters of domain randomization are shown in \tabref{tab:domain_randomization}. We add noise to the proprioceptive observation to make the policy tolerant of the sensor noise in the physical deployment following \cite{rudin2022learning}.
In addition, to make the policy more robust to the impulse, the robot is set to be pushed to 2 m/s from a random direction every 15 seconds.

\section{Experiments and Results} \label{sec:experiments}
In this section, we try to answer the following questions with a series of simulation and physical sim-to-real experiments:
\begin{enumerate}[label=\textbf{Q\arabic*}), leftmargin=*]
    \item Is robustness-based training method enough for dynamic unknown heavy load adaptation task? \label{ques:robustness}
    \item Can the proposed load characteristics modeling method contribute to the load adaptation effectively? \label{ques:load_model_effect}
    \item Can our method reconstruct or infer the load characteristics from proprioceptive sensing with sufficient accuracy to enable load adaptation, thereby achieving unknown dynamic load adaptation on complex terrains? \label{ques:proposed_effect}
\end{enumerate}
We compared the experiment results of our policy and that of the three policies described below. All these three models were trained with the two-stage teacher-student pipeline and an asymmetric actor-critic architecture shown in \figref{fig:training_pipeline} employing the same training configuration detailed in \secref{sec:training_details}. The specifics of them are detailed below and summarized in \tabref{tab:model_train_setting}.
\begin{itemize}
    \item \textbf{Non-load-privileged Walk Model (NLW)}: The privileged observation excludes load characteristics and it is trained without load-related rewards, which refers to \cite{dao2022sim} and represents the robustness-based baseline.
    \item \textbf{Load-privileged Walk Model (LW)}: The load characteristics are available as privileged information to both the critic and the privileged encoder with load-related rewards employed. 
    \item \textbf{Oracle}: 
    The Oracle policy was trained with load characteristics accessible to the actor network, incorporating load-related rewards, and the actor retained access to load characteristics in simulation experiments.
\end{itemize}
\begin{table}[htb!]
    \vspace{-5pt}
    \centering
    \caption{Training Setting for Comparative Experiment}
    \begin{tabular}{lcccc}
    \toprule
         Policy  & NLW & LW & Oracle & \textbf{Ours} \\
    \midrule
         Sim. Env. w/ Load  & \Checkmark & \Checkmark & \Checkmark & \Checkmark  \\
         Load Char. as Priv. Info.  & \XSolidBrush & \Checkmark & \Checkmark & \Checkmark \\
         Load Char. as Actor Input  & \XSolidBrush & \XSolidBrush & \Checkmark & \XSolidBrush \\
         Load Estimator  & \XSolidBrush & \XSolidBrush& \XSolidBrush & \Checkmark \\
         Load Rewards   & \XSolidBrush & \Checkmark & \Checkmark & \Checkmark \\  
    \bottomrule
    \end{tabular}
    \label{tab:model_train_setting}
    \vspace{-10pt}
\end{table}
Both the simulation evaluation and physical validation are divided into dynamic and stationary parts. And the simulation is conducted on the Mujoco simulator\cite{erez2015simulation}.
\begin{figure}[htb!]
    \centering
    \includegraphics[width=0.8\linewidth]{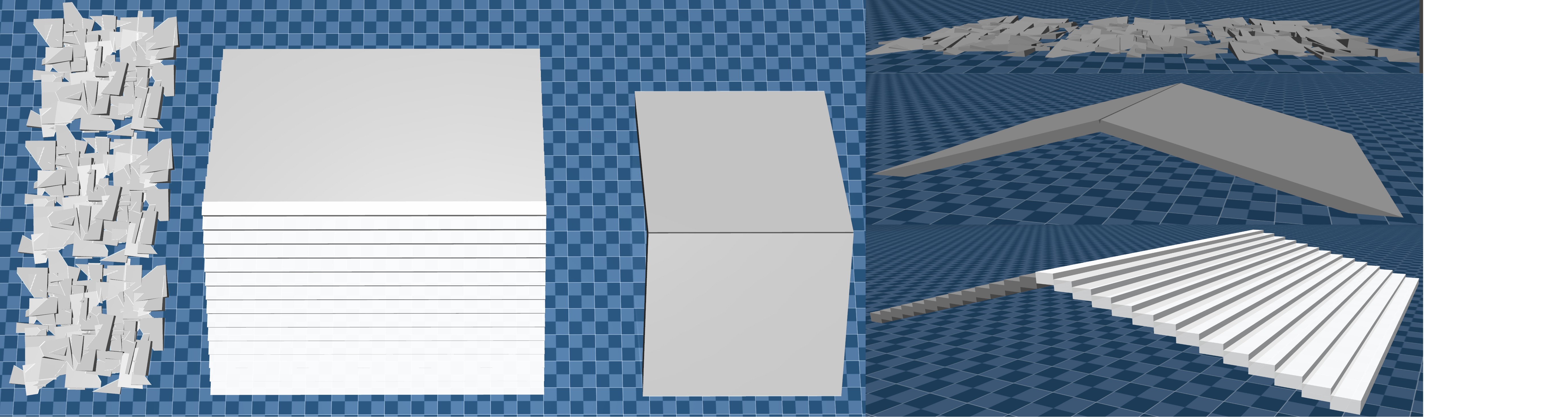}
    \caption{Terrain: rough terrain, stair, slope. The slope has an inclination angle of 26 degrees, and the height and length of each stair step is 0.05 m and 0.2 m. }
    \label{fig:terrain_mujoco}
    \vspace{-10pt}
\end{figure}
\begin{figure*}[htbp]
    \centering
    \includegraphics[width=0.9\linewidth]{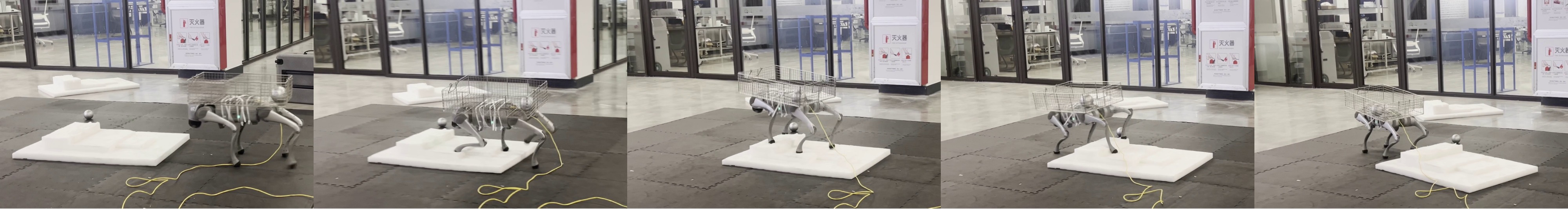}
    \caption{Sim-to-real experiment with 4 kg lead ball as dynamic load on uneven and soft terrain.}
    \label{fig:sim2real_exp}
    \vspace{-10pt}
\end{figure*}

\subsection{Evaluation with Simulation}
\subsubsection{Dynamic Experiment}
\begin{figure}[htbp]
    \centering
    \subfigure[Linear velocity tracking error.]{\includegraphics[width=0.8\linewidth]{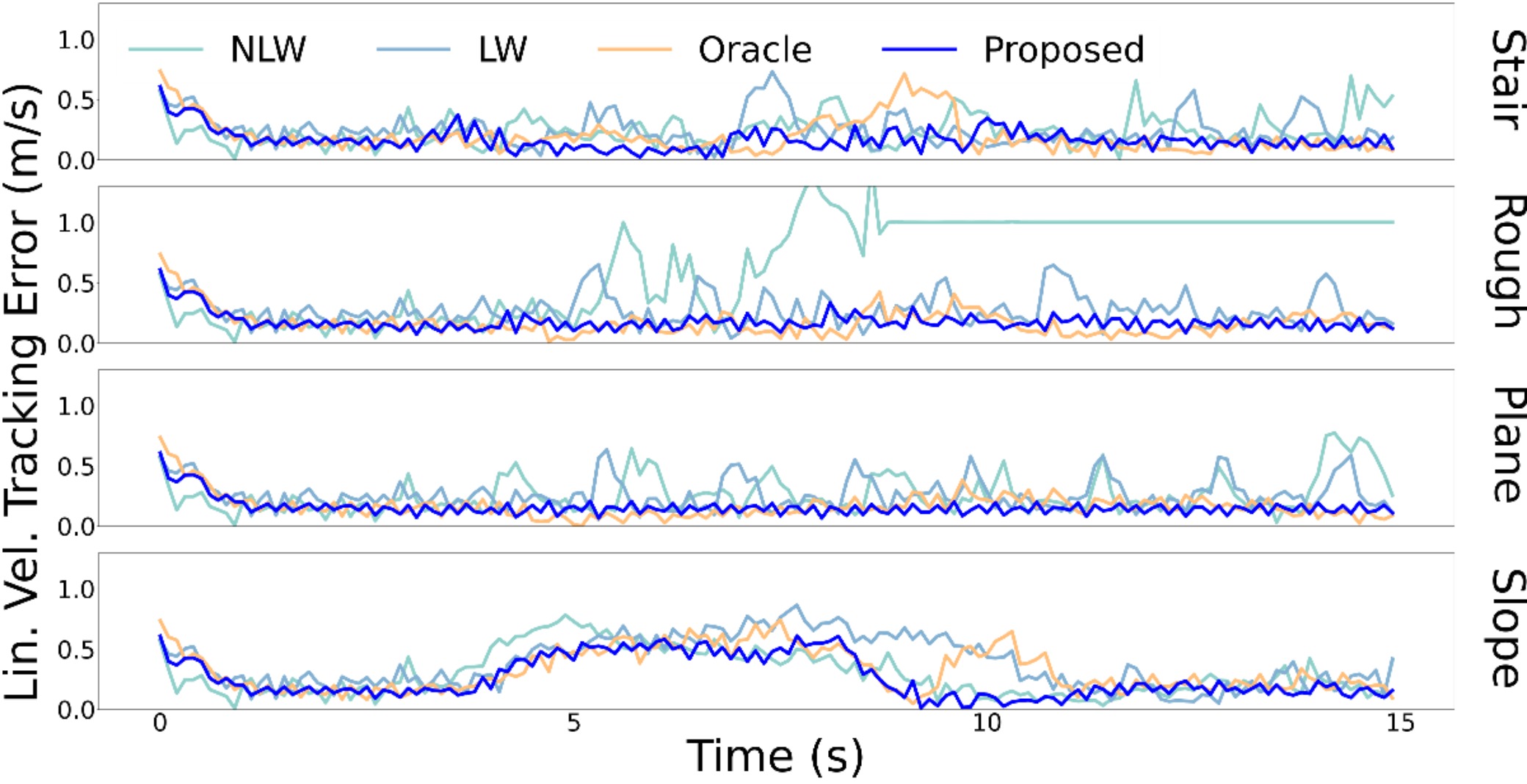}\label{fig:dyna_exp_lin_vel_error}}
    \subfigure[Base roll deviation.]{\includegraphics[width=0.8\linewidth]{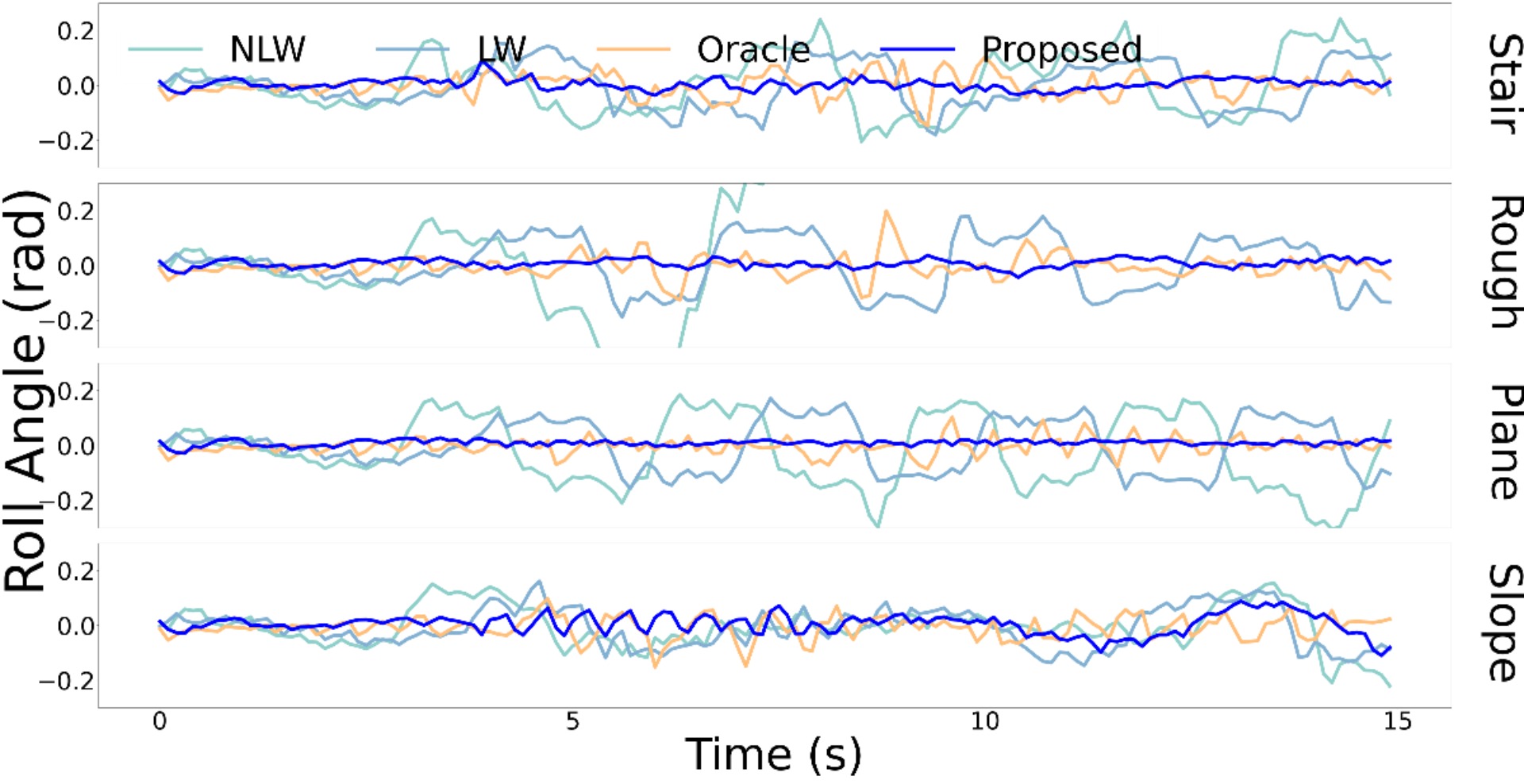}\label{fig:dyna_exp_roll_deviation}}
    \subfigure[Load velocity in robot base frame.]{\includegraphics[width=0.8\linewidth]{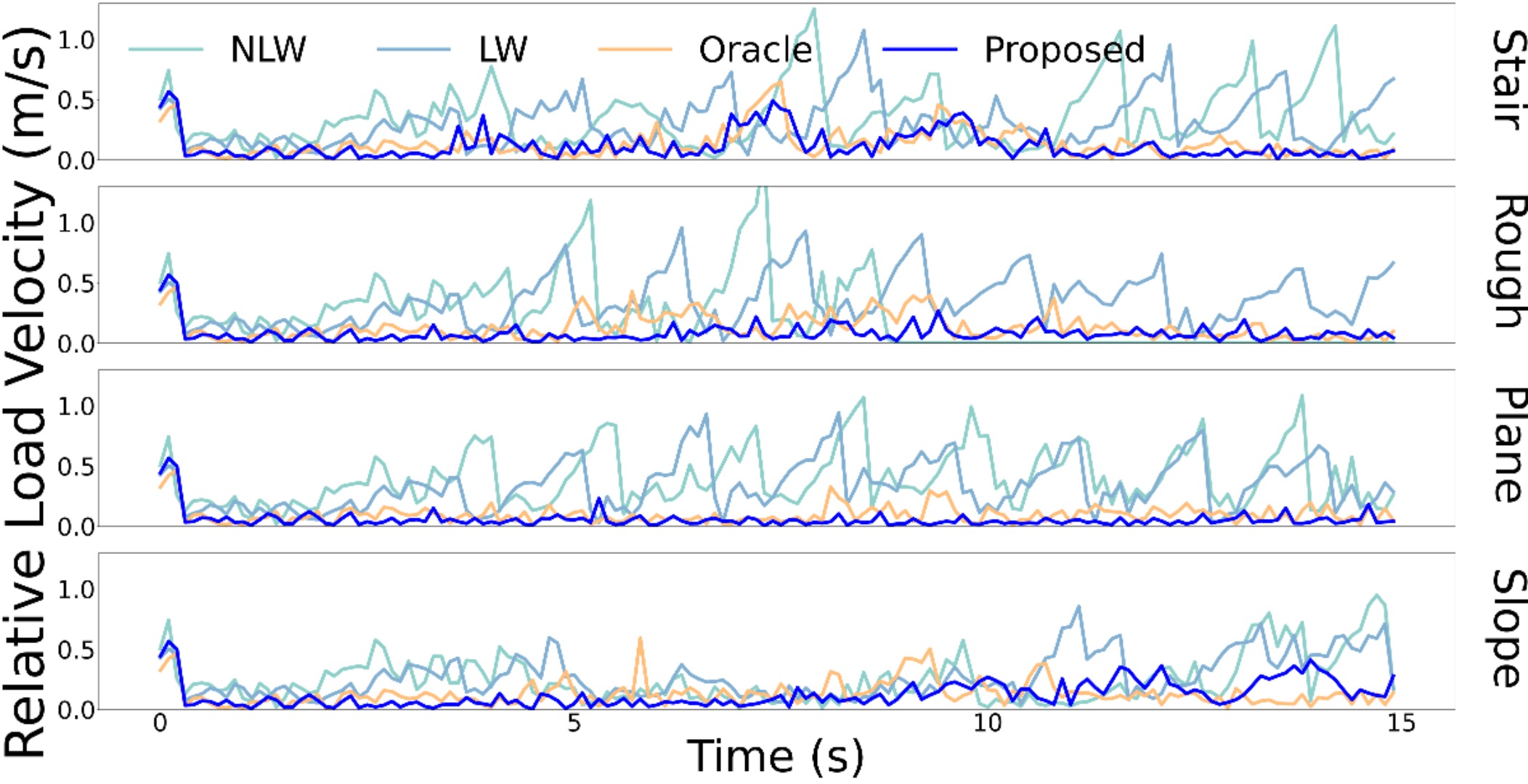}\label{fig:dyna_exp_rel_load_vel}}
    \caption{Results of dynamic experiments under 7 kg load with 0.01 friction coefficient. The robot is commanded to move with 1 m/s velocity along the x-axis (forward). The stair step is 0.05 m high and 0.2 m wide, and the slope is of 26-degree inclination in the pitch direction.}
    \label{fig:dyna_exp}
    \vspace{-15pt}
\end{figure}

We conducted dynamic experiments in 4 scenarios: plane, stair, rough terrain and slope, as shown in \figref{fig:terrain_mujoco}. We collected linear velocity tracking error, robot base roll deviation angle, and load velocity in the robot's base frame for 15 seconds, as displayed in \figref{fig:dyna_exp}.
\figref{fig:dyna_exp_lin_vel_error} demonstrates that the proposed method shows significantly lower velocity tracking error with lower variance compared to NLW and LW in all four types of terrains under heavy and highly dynamic load and achieves performance close to that of the Oracle model. Notably, LW led to the robot falling on the rough terrain, as indicated by the straight line in the second row of \figref{fig:dyna_exp_lin_vel_error}. 
\figref{fig:dyna_exp_roll_deviation} highlights that the proposed method makes the robot base orientation much more stable than the 
NLW and LW and it demonstrates performance on par with the Oracle model. 
\figref{fig:dyna_exp_rel_load_vel} displays the load velocity in the robot base frame, which reveals the model's ability to stabilize the dynamic load. As depicted in \figref{fig:dyna_exp_rel_load_vel}, the load relative velocity data collected of the proposed model displays lower velocity and lower variance compared to NLW and LW, and exhibits a performance level similar to the Oracle model. 

\subsubsection{Stationary Experiment}
\begin{figure}
    \centering
    \subfigure[Robot adaptation of orientation upon load falling.]{\includegraphics[width=0.8\linewidth]{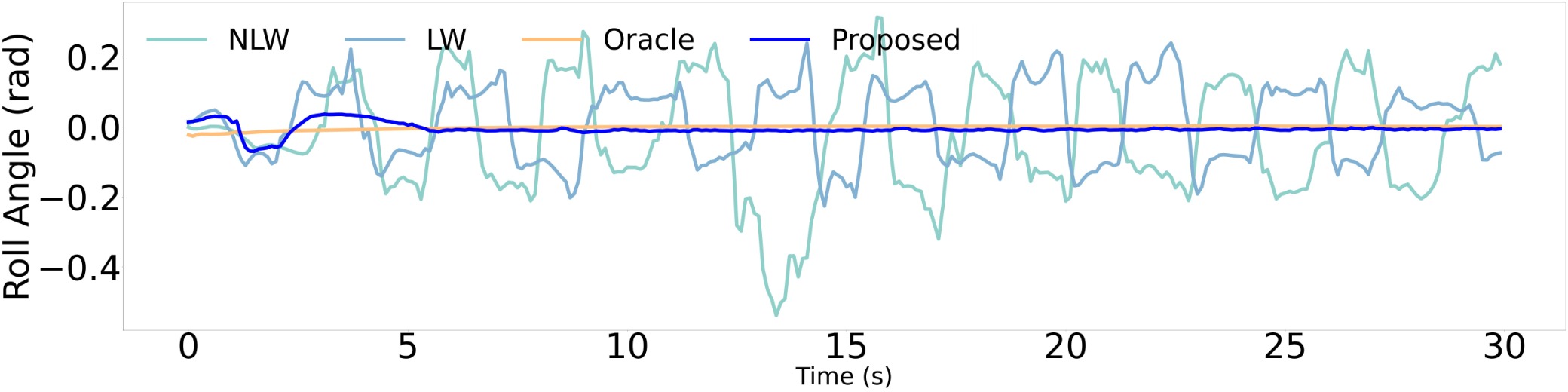}\label{fig:orient_adapt}}
    \subfigure[Load velocity relative to the robot during the adaptation. ]{\includegraphics[width=0.8\linewidth]{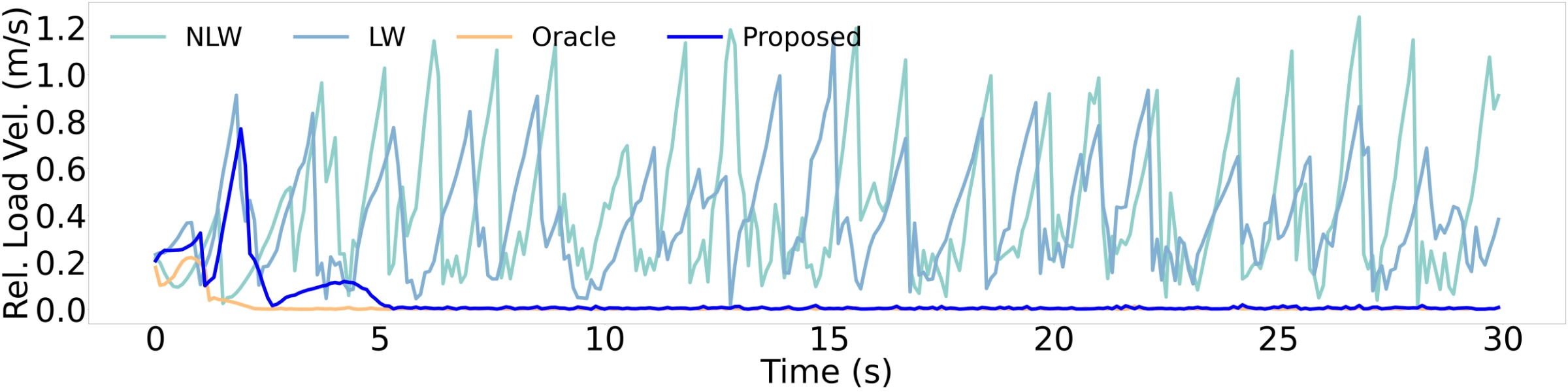}\label{fig:load_vel}}
    \caption{Stationary experiment with 7 kg load and 0.02 friction coefficient.}
    \label{fig:stationary_exp_roll_vel}
    \vspace{-10pt}
\end{figure}

\begin{figure}[htb!]
    \centering\includegraphics[width=0.6\linewidth]{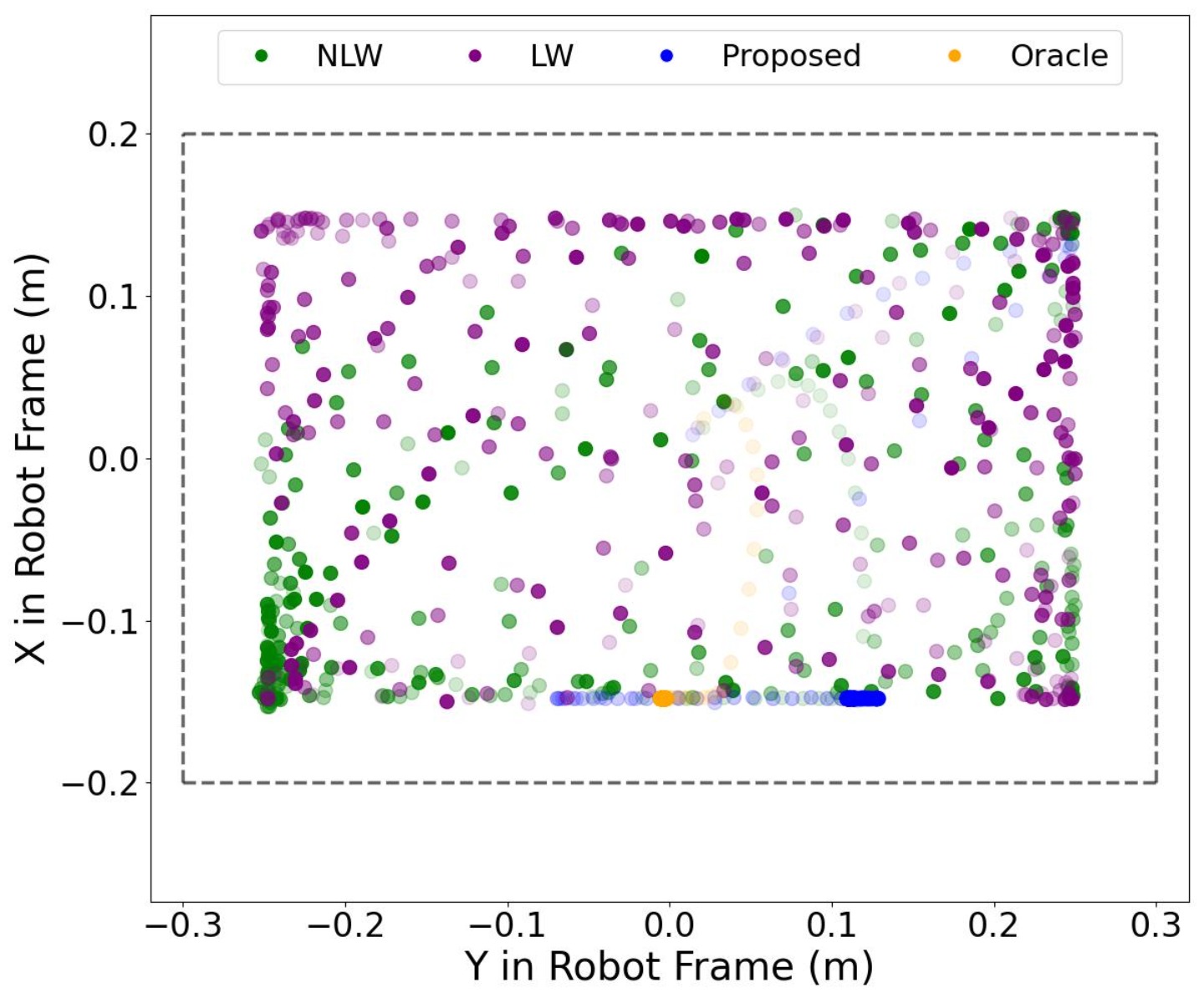}
    \caption{Load trajectory during the adaptation. Higher transparency showing earlier positions, lower transparency showing more recent ones.}
    \label{fig:load_traj}
    \vspace{-15px}
\end{figure}

Stationary experiments are set to evaluate the capability of the robot to handle the load that suddenly appears when standing still. In the experiment, a slippery box with a size of 0.1 m, mass of 7 kg and a friction coefficient of 0.02 appears at 0.3 m above the CoM of the robot base with 0.2 m/s velocity along both 
 x-axis and y-axis. \figref{fig:orient_adapt} displays the robot roll angle variation in the 30 seconds upon the load appearing, where our proposed method demonstrates the shortest settling time while the roll orientation of NLW and LW diverges.
\figref{fig:load_vel} illustrates load relative velocity similar to \figref{fig:dyna_exp_rel_load_vel}, which shows that the proposed method can stabilize the dynamic load after some adaptation motions, exhibiting much better adaptability for the dynamic load. 
\figref{fig:load_traj} illustrates the trajectory of 30 seconds after the load appears, where we can see that the position distribution of the proposed is much denser than that of NLW and LW, indicating the fairly good dynamic load adaptation performance.

Based on the results mentioned above, we can answer the questions posed earlier.
\textbf{To \ref{ques:robustness}:} the robot falling indicated at the second row of \figref{fig:dyna_exp_lin_vel_error} and rapid orientation variation and large load velocity displayed in \figref{fig:dyna_exp}, \figref{fig:orient_adapt} and \figref{fig:load_vel} shows that robustness-based method is not enough for heavy dynamic unknown load adaptation task. 
\textbf{To \ref{ques:load_model_effect}:} the outstanding performance of Oracle model in velocity tracking and load stabilizing revealed by data displayed by \figref{fig:dyna_exp}, \figref{fig:orient_adapt}, \figref{fig:load_vel} and \figref{fig:load_traj} provides strong evidence for the effectiveness of proposed load characteristics model for enabling the robot to capture the load dynamics.
\textbf{To \ref{ques:proposed_effect}:} with only proprioceptive sensing, the proposed method achieves performance close to that of Oracle in terms of velocity tracking on rough terrains, as well as orientation adaptation and load stabilization in both dynamic environments with complex terrains and stationary settings.

\subsection{Physical validation}
We realized the sim-to-real deployment to validate the effectiveness of our proposed method, and the physical experiment scenarios are displayed in \figref{fig:cover_figure} and \figref{fig:sim2real_exp}, where Unitree Go2 quadruped robot succeeded in traversing a soft step carrying a 4 kg dynamic lead ball. We also conducted a series of stationary real-world experiments using 2 kg, 4 kg and 6 kg lead ball that falls from an initial height and with an initial velocity. The real-world experiments show that our proposed method is zero-shot sim-to-real transferable and can enable the robot to adapt itself to the unknown dynamic load with only proprioceptive when both traversing the complex terrain and standing (see supplementary video for details).

\vspace{-0pt}
\section{Conclusion and Future Work} \label{sec:conclusion_future}
In this work, we proposed a framework that enables the quadruped robot to move on rough terrain while carrying an unknown dynamic load. We proposed the load characteristics model for generally representing the load. Utilizing the load characteristics, we developed an RL training framework to train a policy for blind quadrupedal dynamic load carrying.
The evaluation results show that our proposed framework enables the quadruped robot to stabilize the load, keep a balanced orientation and track the velocity command on the rough terrain and outperforms the robustness-based method. And we also conducted sim-to-real deployment and validated its effectiveness in both dynamic and stationary scenes. 
In previous trials, we directly modeled the wrench dynamics from the dynamic load, a lower-dimension and more straight forward quantity for external force handling, but performance was much poorer than that of the proposed method. 
We speculate that this is due to the low-dimensional data exhibiting a better structure in the high-dimensional space, making it easier for the policy to capture the underlying data distribution, which could be investigated more in the future work. In addition, our proposed control strategy did not show a clear advantage over the baseline on sloped terrain according to the results in \figref{fig:dyna_exp}, which could be a direction for further improvement.

\flushend 

\clearpage
\printbibliography

\end{document}